\documentclass[letterpaper, 10 pt, conference]{ieeeconf}  

\IEEEoverridecommandlockouts                              

\usepackage{graphics} 

\usepackage{algorithm}
\usepackage{algorithmic}
\usepackage{amsmath} 
\usepackage{amssymb}
\usepackage{arydshln}
\usepackage{newfloat}
\usepackage{listings}

\usepackage{booktabs}      
\usepackage{multirow}      
\usepackage[table]{xcolor} 
\usepackage{graphicx}      
\usepackage{xcolor}

\title{\LARGE \bf
DiffVL: Diffusion-Based Visual Localization on 2D Maps via BEV-Conditioned GPS Denoising
}

\author{Li Gao$^{1 \ast}$ Hongyang Sun$^{2 \ast \dag}$ Liu Liu$^{1 \ddag}$ Yunhao Li$^{1 \dag}$ Yang Cai$^{1}$ \\
\emph{\{liangliang.gl, diana.ll, huaimu.lyh\}@alibaba-inc.com} \\
\emph{sunhongyang10@zju.edu.cn} \\
\emph{yangcai.cy@autonavi.com} \\
\thanks{* Equal Contribution}
\thanks{\dag Work done during the internship at Amap, Alibaba Group}
\thanks{\ddag Corresponding Author}
\thanks{$^{1}$AMAP, Alibaba Group}
\thanks{$^{2}$Zhejiang University}
}

\begin{document}

\maketitle
\thispagestyle{empty}
\pagestyle{empty}

\begin{abstract}
Accurate visual localization is crucial for autonomous driving, yet existing methods face a fundamental dilemma: While high-definition (HD) maps provide high-precision localization references, their costly construction and maintenance hinder scalability, which drives research toward standard-definition (SD) maps like OpenStreetMap. Current SD-map-based approaches primarily focus on Bird's-Eye View (BEV) matching between images and maps, overlooking a ubiquitous signal-noisy GPS. Although GPS is readily available, it suffers from multipath errors in urban environments.
We propose DiffVL, the first framework to reformulate visual localization as a GPS denoising task using diffusion models. Our key insight is that noisy GPS trajectory, when conditioned on visual BEV features and SD maps, implicitly encode the true pose distribution, which can be recovered through iterative diffusion refinement. DiffVL, unlike prior BEV-matching methods (e.g., OrienterNet) or transformer-based registration approaches, learns to reverse GPS noise perturbations by jointly modeling GPS, SD map, and visual signals, achieving sub-meter accuracy without relying on HD maps.
Experiments on multiple datasets demonstrate that our method achieves state-of-the-art accuracy compared to BEV-matching baselines. Crucially, our work proves that diffusion models can enable scalable localization by treating noisy GPS as a generative prior—making a paradigm shift from traditional matching-based methods. Code and models will be open-sourced.

\end{abstract}

\section{Introduction}

Visual localization is a critical technology for applications like autonomous driving \cite{yaqoob2019autonomous,levinson2011towards}, augmented reality, and robotics, where precise and reliable pose estimation is paramount for safe navigation \cite{cai2025navdp} and decision-making \cite{liao2025diffusiondrive}. The core task involves estimating a 3-DoF pose (position and orientation) from visual imagery against a 2D map. To meet the stringent demands of autonomous systems, traditional methods \cite{barsan2020learning} have heavily relied on High-Definition (HD) maps. However, the high costs of creating, annotating, and frequently updating these maps severely limit their scalability and widespread adoption, creating a significant bottleneck for deploying autonomous technology at a global scale.

In response, recent research has shifted towards localization with low-cost, globally available Standard-Definition (SD) maps, such as OpenStreetMap \cite{haklay2008openstreetmap}. Representative works \cite{sarlin2023orienternet, zhou2025seglocnet} leverage deep learning \cite{gao2021dsp, gao2021addressing,gao2022doubly,xing2022bsam} to infer the 3-DoF pose by aligning a Bird's-Eye-View (BEV) representation, derived from the input image, with an SD map. While these approaches have shown promise, they are susceptible to challenges like perceptual aliasing in visually repetitive areas and typically overlook a crucial, ubiquitous source of information: noisy GPS data. This omission inherently limits the upper bound of their localization accuracy and robustness, particularly in challenging scenarios.

\begin{figure}[t]
    \centering
    \includegraphics[width=\linewidth]{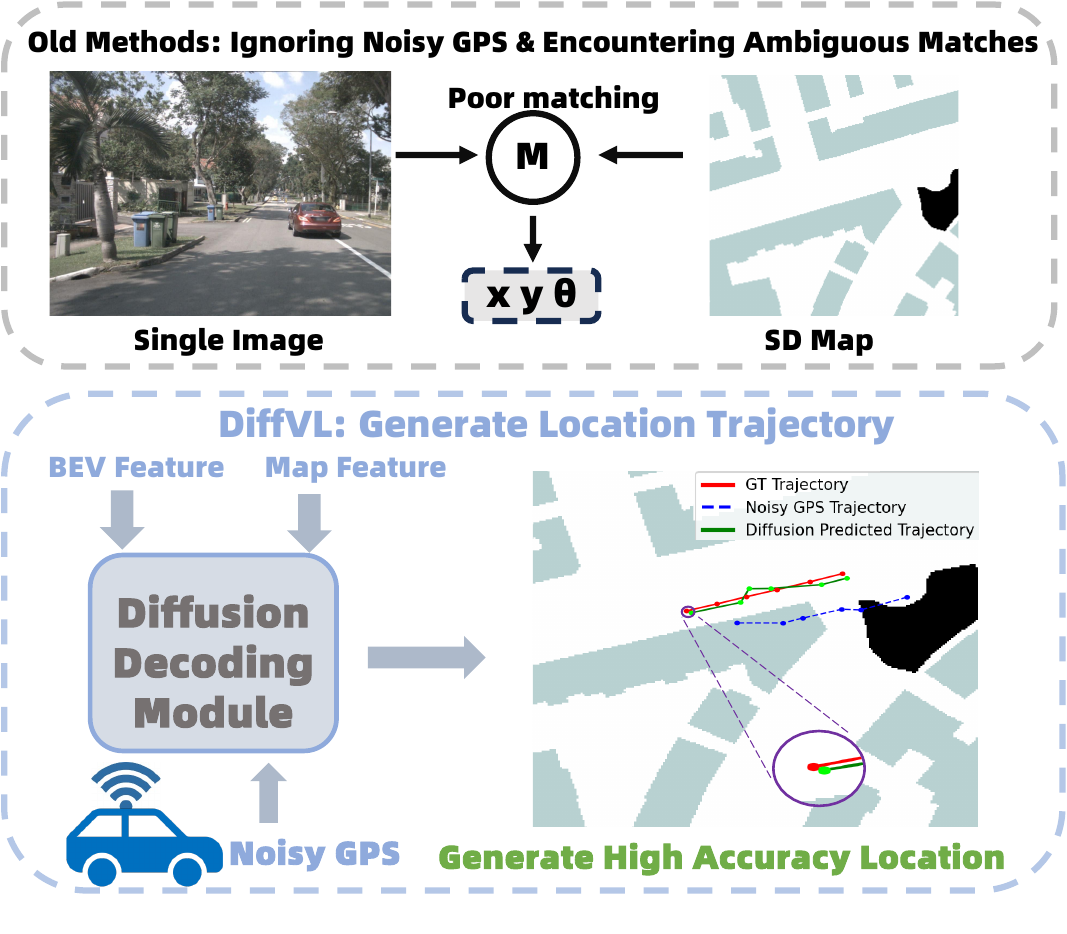}
    \caption{Overview of the proposed DiffVL. Most existing SD map-based localization methods rely on exhaustive geometric matching between Bird's-Eye View (BEV) features and map elements to compute the pose. In contrast, our approach fundamentally reformulates visual localization as a generative modeling task.}
    \label{fig:teaser}
\end{figure}
In recent years, diffusion models \cite{ho2020denoising, song2020denoising,jiang2024scenediffuser,croitoru2023diffusion,li2024generative}, have emerged as powerful generative techniques, achieving breakthrough performance by learning to reverse gradual noising processes. Such techniques have been successfully applied in embodied navigation\cite{trackvla} and autonomous driving\cite{transdiffuser} for predicting future trajectories from motion anchors. Inspired by these advances, we present a paradigm shift for visual localization: we reformulate it as a conditional denoising process of GPS trajectories. Our key insight is that noisy GPS signals, often dismissed as unreliable, actually encode the true pose distribution and can be transformed into precise localization through diffusion-based denoising. Specifically, we reframe visual localization from a traditional image-to-map matching task \cite{quddus2007current,pink2008visual,huang2021survey} into a conditional generation problem, where a diffusion model learns to reverse the noise corruption in GPS trajectories conditioned on visual observations.

We propose DiffVL, a novel diffusion-based framework that synergistically integrates sequential GPS signals and visual cues. To imbue the diffusion model with geometric and semantic awareness, we introduce a dual-objective training strategy. The primary trajectory refinement loss ($\mathcal{L}_{\text{diff}}$) ensures kinematic and temporal coherence in the denoised trajectory. Simultaneously, an auxiliary localization prior loss ($\mathcal{L}_{\text{loc}}$), computed by aligning Bird's-Eye View (BEV) visual features with map elements, provides strong geometric regularization. This dual loss compels shared feature encoders to learn representations that are both visually discriminative and geometrically consistent with the map's coordinate system. Through joint optimization, our model achieves a robust balance between motion-based prediction and appearance-based matching, enabling high-precision pose estimation from noisy GPS and significantly enhancing localization robustness.

Our main contributions are summarized as follows:

\begin{itemize}
    \item We introduce DiffVL, a novel visual localization paradigm that, to the best of our knowledge, is the first to successfully and principally apply diffusion models to denoise noisy GPS trajectories for this task. It reframes localization from a matching problem to  conditional generation problem.
    \item We redefine the role of noisy GPS signals in visual localization. Whereas prior research often ignored or filtered these signals, we are the first to frame them as a `Noisy Observation' of the true pose distribution, providing a methodological basis for leveraging generative models to recover high-precision poses.
    \item We design a dual-objective training framework where a trajectory refinement loss and a localization prior loss work in tandem. This ensures the model not only recovers a coherent trajectory but also learns a powerful, geometrically consistent feature representation from visual and map data.
    \item We conduct rigorous and comprehensive evaluations on multiple large-scale autonomous driving datasets (KITTI\cite{kitti}, nuScenes\cite{caesar2020nuscenes}, and MGL\cite{sarlin2023orienternet}). Our experimental results demonstrate that DiffVL significantly outperforms existing methods across all datasets, achieving state-of-the-art performance. We plan to release our code and models to facilitate future research and contribute to the community.
\end{itemize}
\section{Related Work}

\subsection{Map-based Visual Localization}

Map-based visual localization is a long-standing research task central to robotics \cite{cai2025navdp} and autonomous systems\cite{li2019aads, hu2023planning, yan2024street}. Traditional methods rely on matching sensor data against meticulously pre-built 3D maps\cite{li2012worldwide, shi2022beyond}. These maps are typically composed of dense point clouds from LiDAR sensors or are reconstructed via Structure-from-Motion (SfM) from multiple image views \cite{agarwal2011building}. They enable high-precision pose estimation through the alignment of handcrafted or learned local features between the query image and the 3D model\cite{lowe2004distinctive, bay2006surf}, or through direct geometric registration of point clouds, often using variants of the Iterative Closest Point (ICP) algorithm \cite{zhang2021fast}. However, these High-Definition (HD) maps come with significant practical drawbacks. Their creation requires specialized vehicles and extensive surveying, their maintenance is a continuous and costly effort, and their large memory footprint poses a major barrier to on-board storage and large-scale deployment.

To overcome these scalability issues, recent research has shifted towards Standard-Definition (SD) maps. These maps, often derived from crowd-sourced data like OpenStreetMap (OSM) \cite{haklay2008openstreetmap}, are lightweight, globally-available, and semantically rich, making them a highly attractive alternative. A prominent line of work involves generating a neural Bird's-Eye-View (BEV)\cite{zhang2025bev, li2024bevformer} representation from a monocular camera and then aligning it with a corresponding rasterized map tile \cite{sarlin2023orienternet, zhou2025seglocnet}. These methods effectively tackle the cross-view matching problem between a ground-level perspective and a top-down map. Despite their promising performance, these approaches face two key challenges. First, they must bridge the significant domain gap between rendered map semantics and complex, real-world visual features. Second, they almost universally overlook an important and readily accessible signal: noisy GPS data. This omission of a critical information source, often due to the difficulty of handling its inherent noise, inherently limits their robustness and accuracy, particularly in ambiguous environments.

\subsection{Diffusion Models}

In recent years, Denoising Diffusion Probabilistic Models (DDPMs) \cite{ho2020denoising, song2020denoising, lipman2022flow} have emerged as a disruptive technology in generative AI, achieving state-of-the-art results in diverse domains such as high-fidelity image and video generation \cite{zhang2023text, yan2025streetcrafter}, robotic policy learning \cite{chi2023diffusion}, and motion prediction \cite{jiang2023motiondiffuser}. The core idea is to learn to reverse a fixed Markov chain that gradually adds noise to data, allowing the model to generate new samples by progressively denoising from pure Gaussian noise. Their ability to capture complex, multi-modal distributions makes them particularly well-suited for robotics and autonomous systems. For instance, Diffusion Policy \cite{chi2023diffusion, shipan2008mechanisms} has proven highly effective at learning multi-modal action distributions for complex manipulation tasks. In motion prediction, works like MotionDiffuser \cite{jiang2023motiondiffuser, barquero2023belfusion} leverage conditional diffusion models to generate kinematically coherent and socially-aware multi-agent trajectories.

Despite their demonstrated potential, the application of diffusion models to visual localization remains an unexplored frontier. Inspired by these pioneering works\cite{liao2025diffusiondrive, wang2025diffad, zhao2025diffe2e}, we are the first to introduce the diffusion model paradigm to this task. Our key insight is to reframe the problem: instead of treating localization as a deterministic matching problem\cite{sarlin2023orienternet, zhou2025seglocnet}, we embrace the inherent uncertainty of sensor data by formulating it as a conditional denoising task. In our DiffVL framework, the diffusion model learns to recover a kinematically smooth trajectory from a noisy raw GPS sequence. Crucially, this denoising process is not performed in isolation; it is intelligently conditioned on rich, multi-modal features extracted from both the camera image and the SD map\cite{haklay2008openstreetmap}, providing essential environmental context. To ensure these conditional features are geometrically grounded and semantically meaningful, we employ a dual-objective training strategy. A primary trajectory refinement loss is complemented by an auxiliary BEV-map matching loss, which acts as a powerful geometric regularizer, forcing the shared encoders to learn a spatially-aware representation. This approach directly integrates the generative power of diffusion models with the discriminative task of visual-map matching at the feature level, significantly enhancing both localization robustness and accuracy.

\begin{figure*}[t]
    \centering
    \includegraphics[width=\linewidth]{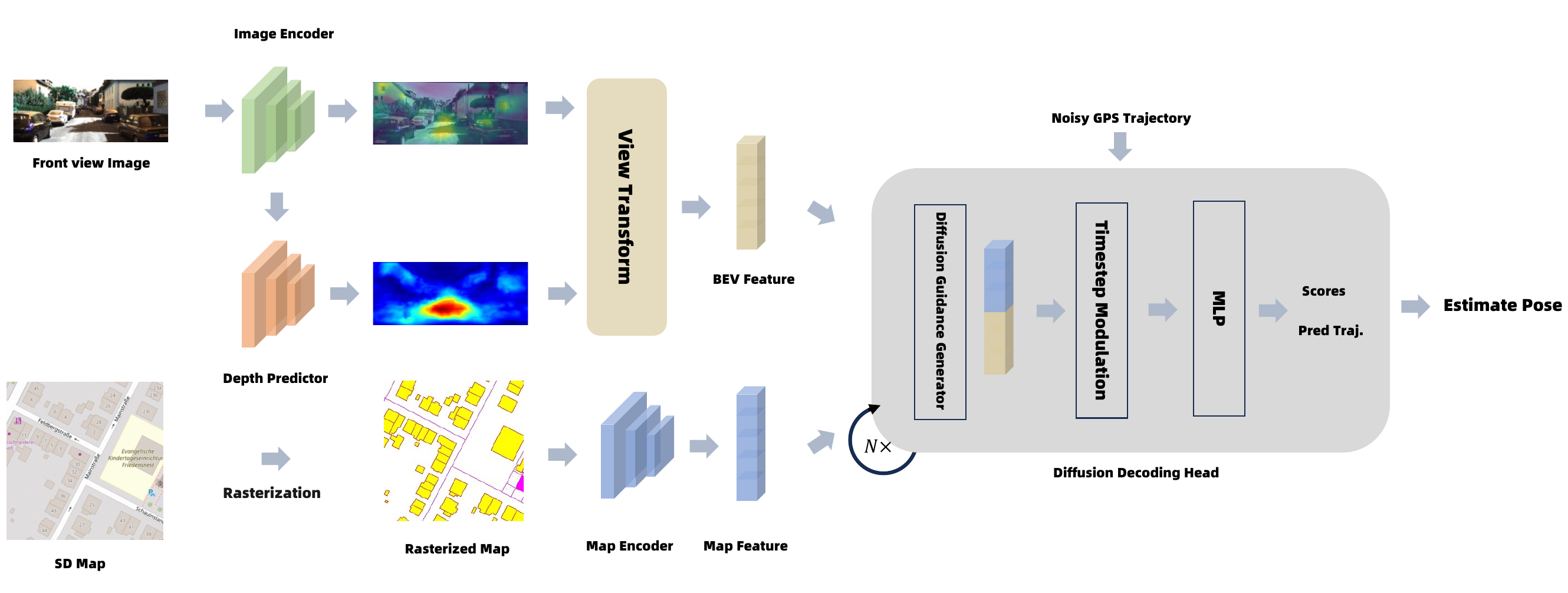}
    \caption{Architecture of DiffVL. As the first visual localization framework built upon diffusion models, our system pioneers a paradigm shift from traditional matching-based approaches to a generative formulation. The architecture accepts three critical inputs: (i) a monocular front-view RGB image capturing immediate scene context, (ii) standard-definition (SD) map data providing structural priors, and (iii) a noisy GPS trajectory offering coarse positional cues. Central to our innovation, the Image Encoding Module transforms perspective views into geometrically consistent Bird's-Eye-View (BEV) features, while the Map Encoding Module extracts topological representations from SDmaps. These complementary features undergo multi-modal fusion to generate conditioning features for our novel diffusion module—the core component that fundamentally redefines visual localization as a conditional generation task. Through iterative reverse diffusion steps, this module progressively denoises the corrupted GPS input, transforming unreliable sensor measurements into precise 3-DoF pose estimates. This generative approach marks the first successful application of diffusion models to visual localization, establishing a new trajectory refinement paradigm.}
    \label{framework}
\end{figure*}

\begin{figure*}[t]
    \centering
    \includegraphics[width=0.86\linewidth]{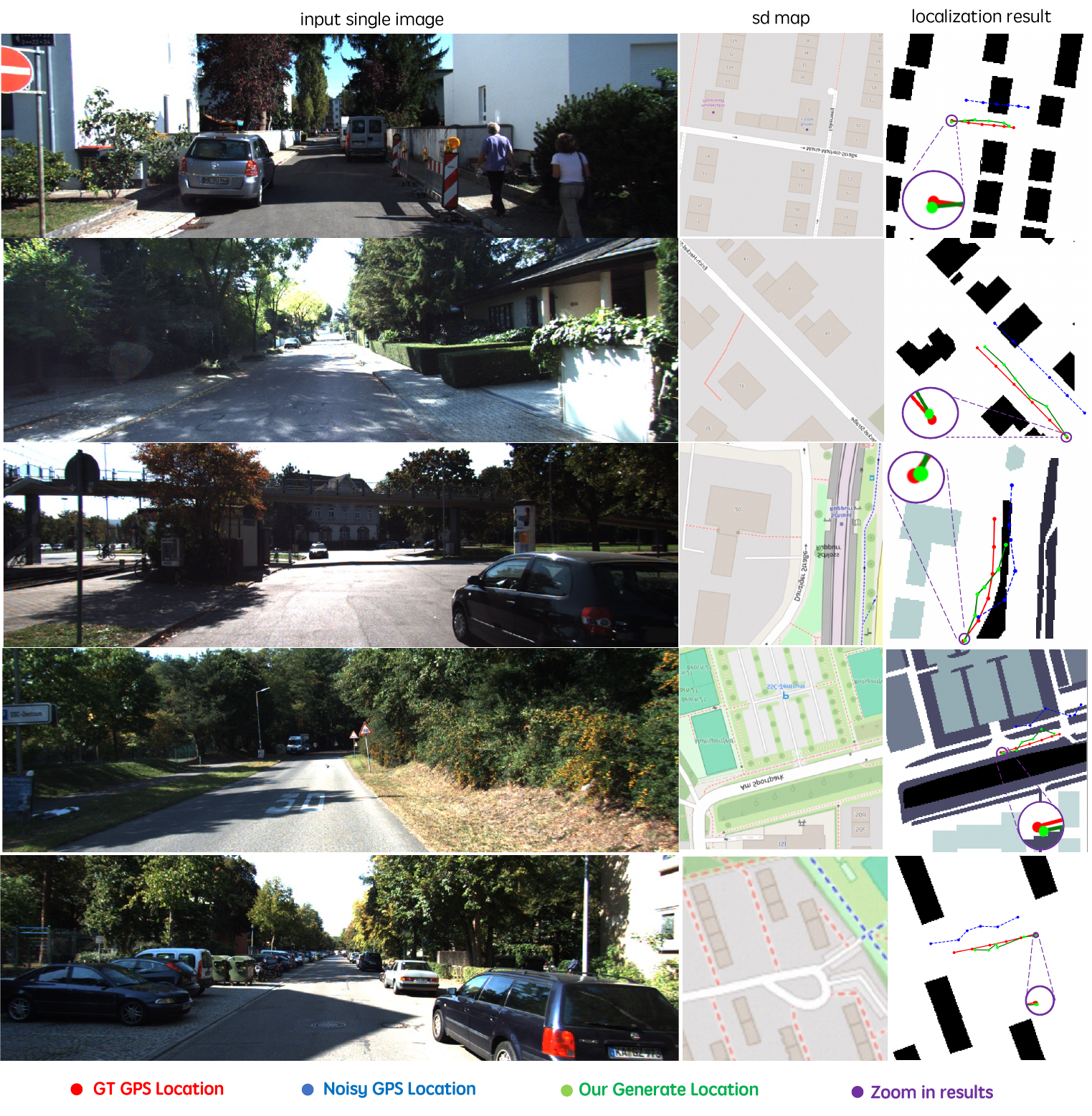}
    \caption{The localization results of our method on the KITTI dataset. In these visualizations, the red trajectory represents the ground truth (GT) GPS trajectory from the dataset, while the blue trajectory is the noisy GPS trajectory we synthetically generate. Given the noisy blue trajectory and a single image as input, our method produces the refined green "Generated Location" trajectory.}
    \label{visualize}
\end{figure*}

\section{Method}
\subsection{Problem Formulation}
Given a single front-view image $\mathcal{I}$ and a historical sequence of noisy GPS measurements $\mathbf{P}^{\text{gps}} = \left\{ \mathbf{p}^{\text{gps}}_t \right\}_{t=1}^T = \left\{ x^{\text{gps}}_t, y^{\text{gps}}_t \right\}_{t=1}^T$ where $T$ is the time horizon and $ \mathbf{p}^{\text{gps}}_t=(x_t^{\text{gps}}, y_t^{\text{gps}})$ denotes the East-North-Up (ENU) coordinates at timestep $t$, along with an SD map $M$ representing the local map tile encompassing the trajectory $\mathbf{P}^{\text{gps}}$. Our task is to estimate a 3-DoF pose $\hat{\mathbf{p}} = (x, y, \theta) \in \mathbb{R}^3$, where $(x, y)$ represents the ENU position and $\theta \in (-\pi, \pi]$ denotes the heading angle around the vertical $\mathbf{z}$-axis.

The pose estimation process is formalized as a conditional diffusion model:
\begin{equation}
\hat{\mathbf{p}} = \mathcal{A}_{\theta}\left (\left \{  \mathbf{p}^{\text{gps}}_t \right \}_{i=1}^{T} \mid \mathbf{z}   \right)
\end{equation}
where:
\begin{itemize}
    \item $\mathcal{A}_{\boldsymbol{\theta}}(\cdot)$ denotes the diffusion model head
    \item $\boldsymbol{\theta}$ represents trainable parameters
    \item $\mathbf{z} = f(\mathcal{I}, M)$ is the conditional latent vector encoding image-map features
\end{itemize}

\subsection{Overview}
The DiffVL framework comprises four core modules with the following architecture, as shown in figure \ref{framework}.

\textbf{Image Encoder}:
Performs environmental perception through front-view images, extracting multi-scale visual features. Utilizes depth estimation to analyze scene geometry, combined with view transformation that convert perspective representations into geometrically consistent BEV(bird's-eye view) feature maps. This process effectively preserves spatial relationships and semantic content while establishing correspondence with real-world coordinate systems, enabling robust understanding of the surroundings from an overhead perspective.

\textbf{Map Encoder}:
Leverages OpenStreetMap \cite{haklay2008openstreetmap} to construct rasterized map representations through grid-based processing, encoding critical prior knowledge about the environment.

\textbf{Diffusion Guidance Generator}:
Implements cross-modal fusion between environmental perception and prior map knowledge through attention mechanisms. Generates conditional guidance embeddings that integrate visual observations with map semantics, constructing global contextual representations that guide subsequent diffusion denoising processes.

\textbf{Diffusion Head}:
Achieves probabilistic localization refinement through iterative diffusion-based correction. Formulates pose estimation as a progressive noise reduction problem, where initial position hypotheses undergo multi-stage optimization guided by multimodal contextual features. By jointly optimizing positional and orientation parameters, it systematically reduces localization uncertainty, ultimately achieving meter-level accuracy on the Standard Definition (SD) Map.

Detailed parameter configurations of the system architecture and multi-task loss function designs will be comprehensively presented in subsequent chapters.

\subsection{Image Encoding Module}
This module extracts structured environmental features from a single front-view image $\mathcal{I} \in \mathbb{R}^{H \times W \times 3}$ and transforms them into a Bird's-Eye View (BEV) representation. The implementation consists of three key stages:

\textbf{Multi-scale Feature Extraction}: A ResNet-101\cite{resnet} backbone network is employed to extract multi-scale feature pyramids in the perspective view (PV):
\begin{equation}
\left\{ \mathbf{F}_{\text{pv}}^1, \mathbf{F}_{\text{pv}}^2, \mathbf{F}_{\text{pv}}^3, \mathbf{F}_{\text{pv}}^4 \right\} = \Phi_{\text{img}}(\mathcal{I})
\end{equation}
where $\mathbf{F}_{\text{pv}}^i \in \mathbb{R}^{H_i \times W_i \times C_i}$ denotes the feature map at the $i$-th level. Multi-resolution information is fused across different layers via skip connections, enhancing the model's ability to perceive complex scenes.

\textbf{Depth Probability Distribution Prediction}: To improve geometric modeling accuracy, a parallel depth estimation branch is introduced to predict per-pixel depth distributions from the feature pyramid:
\begin{equation}
\mathcal{D} = \Psi_{\text{depth}} \left( \bigoplus_{i=1}^{4} \text{UpSample}(\mathbf{F}_{\text{pv}}^i) \right)
\end{equation}
Here, $\mathcal{D} \in \mathbb{R}^{H \times W \times D}$ is the depth distribution tensor, where $D$ denotes the number of discrete depth bins, $\Psi_{\text{depth}}$ is the depth prediction subnetwork, and $\bigoplus$ represents feature concatenation.

\textbf{Differentiable View Transformation}: Employs a polar-Cartesian dual projection like OrienterNet\cite{sarlin2023orienternet} and LSS\cite{lss}:
\begin{equation}
\mathbf{F}_{\text{bev}} = \mathcal{P}_{\text{cart}} \left( \mathcal{P}_{\text{polar}} \left( \mathbf{F}_{\text{pv}}, \mathcal{D} \right), \mathcal{C} \right)
\end{equation}
where:
\begin{itemize}
  \item $\mathcal{P}_{\text{polar}}$: Polar coordinate projection with scale priors $\mathcal{D}$
  \item $\mathcal{P}_{\text{cart}}$: Polar-to-Cartesian coordinate transformation
  \item $\mathcal{C}$: Camera intrinsic/extrinsic parameters
\end{itemize}
yielding BEV features $\mathbf{F}_{\text{bev}} \in \mathbb{R}^{B \times C \times H_b \times W_b}$.

\subsection{Map Encoding Module}
This module constructs structured environmental priors from open-source map data through four processing stages:

\textbf{Map Data Acquisition}: A bounding box is computed based on the spatial distribution of historical GPS trajectory $\mathbf{P}^{\text{gps}}$, and the corresponding vector map data for the region of interest is retrieved from OpenStreetMap (OSM). This ensures spatial consistency between the map information and the vehicle's current location, forming the foundation for global prior construction.

\textbf{Semantic Rasterization}: The vector map data is converted into a three-channel RGB image:
Channel 1 encodes the road network (including highways, arterials, and local roads), channel 2 represents building footprints, and channel 3 encodes natural features (such as vegetation, water bodies, and terrain). The rasterization resolution aligns with that of the BEV features ($0.5\text{m}/\text{pixel}$), resulting in a map image $\mathbf{M}_{\text{rgb}} \in \mathbb{R}^{X \times Y \times 3}$. This approach is inspired by previous modeling techniques in \cite{sarlin2023orienternet}, with the objective of improving the system's understanding of static environmental structures.

\textbf{Hierarchical Feature Extraction}: A VGG16\cite{vgg} architecture is used to extract features from the rasterized map:
\begin{equation}
\mathbf{F}_{\text{map}} = \Psi_{\text{MapEnc}}(\mathbf{M}_{\text{rgb}})
\end{equation}
The resulting $\mathbf{F}_{\text{map}}$ is a compressed feature map that captures key priors such as road topology and traversability constraints that emphasize effective representation learning of structured map information.

\subsection{Diffusion Guidance Generator}
This module achieves deep fusion of visual perception and map priors to generate multi-scale contextual features for the diffusion model, comprising of multimodal feature fusion and diffusion decoding head:

\textbf{Multimodal Feature Fusion}: 
The BEV and map features undergo dimensional alignment and contextual integration:
\begin{equation}
\mathbf{F}_{\text{cond}} = \Gamma \left( \phi_{\text{bev}}(\mathbf{F}_{\text{bev}}), \psi_{\text{map}}(\mathbf{F}_{\text{map}}) \right)
\end{equation}
where:
\begin{itemize}
  \item $\psi_{\text{map}}$: Map feature compression and spatial restructuring
  \item $\phi_{\text{bev}}$: BEV feature projection and spatial alignment
  \item $\Gamma$: Cross-modal fusion operator
\end{itemize}
yielding a unified representation $\mathbf{F}_{\text{cond}} \in \mathbb{R}^{B \times C_f \times H \times W}$ that synthesizes visual perception and semantic priors.

\textbf{Diffusion Decoding Head}: This module implements conditional trajectory generation and pose refinement through diffusion modeling. Firstly, historical trajectories are normalized and noise-injected through a linear diffusion process. We normalizes historical GPS trajectories $\mathbf{P}^{\text{gps}} \in \mathbb{R}^{T \times 3}$ to $[-1,1]$ range:
\begin{equation}
\mathbf{P}_{\text{norm}} = \text{norm}_{\text{odo}}(\mathbf{P}^{\text{gps}})
\end{equation}
then injects Gaussian noise:
\begin{equation}
\mathbf{P}_t = \sqrt{\bar{\alpha}_t} \mathbf{P}_{\text{norm}} + \sqrt{1-\bar{\alpha}_t}\epsilon,\ \ \epsilon \sim \mathcal{N}(0,\mathbf{I})
\end{equation}

Inspired by DiffusionDrive\cite{diffusiondrive}, during training, the diffusion decoder takes as input $N_{\text{anchor}}$ noisy trajectories $\left \{  \mathbf{p}^{\text{gps}}_t \right \}_{k=1}^{\text{anchor}}$ and predicts classification scores $\{\hat{s}_k\}_{k=1}^{N_{\text{anchor}}}$ and denoised trajectories $\hat{\mathbf{p}}_{k=1}^{N_{\text{anchor}}}$:
\begin{equation}
\{\hat{s}_k, \hat{\mathbf{p}}_k\}_{k=1}^{N_{\text{anchor}}} = f_\theta \left( \{\mathbf{p}^k\}_{k=1}^{N_{\text{anchor}}}, \mathbf{F}_{\text{cond}} \right)
\end{equation}
where $\mathbf{F}_{\text{cond}}$ represents the conditional information. We assign the noisy trajectory around the closest anchor to the ground truth trajectory $\tau_{\text{gt}}$ as positive sample ($y_k = 1$) and others as negative samples ($y_k = 0$).

\subsection{Loss Formulation}
The total training objective combines trajectory refinement and localization losses:
\begin{equation}
\mathcal{L}_{\text{total}} = \underbrace{\mathcal{L}_{\text{diff}}}_{\text{Trajectory Refinement}} + \alpha \underbrace{\mathcal{L}_{\text{loc}}}_{\text{Localization Prior}}
\end{equation}
where $\alpha$ balances the contribution of the localization loss.

\subsubsection{Trajectory Refinement Loss}
Guides diffusion-based trajectory generation through multi-modal optimization:
\begin{equation}
\mathcal{L}_{\text{diff}} = \sum_{k=1}^{N_{\text{anchor}}} \left[ y_k \left\lVert \hat{\mathbf{p}}_k - \tau_{\text{gt}} \right\rVert_1 + \lambda \mathcal{L}_{\text{BCE}}(\hat{s}_k, y_k) \right]
\end{equation}
with $y_k = \mathbb{I}\left[ k = \underset{j}{\text{argmin}} \left\lVert \mathbf{p}_k  - \tau_{\text{gt}} \right\rVert_2 \right]$, enforcing mode selection around ground truth anchors.

\subsubsection{Localization Loss}
Provides spatial regularization through BEV-map matching:
\begin{align}
&\mathcal{L}_{\text{loc}} = -\log \mathbf{P}(\tau_{\text{gt}} \mid \mathbf{S}) \\
&\mathbf{S} = \text{Match}(\mathbf{F}_{\text{bev}}, \mathbf{F}_{\text{map}})
\end{align}
implemented via label-smoothed negative log-likelihood that accounts for annotation uncertainties.
This multi-objective design enables \textit{Precise trajectory generation} via $\mathcal{L}_{\text{diff}}$ and \textit{Geometric consistency} via $\mathcal{L}_{\text{loc}}$.

\section{Experiment}

\begin{table*}[htbp]
    \centering
    \caption{\textbf{Quantitative comparison on the KITTI dataset.} All metrics are Recall (\%), where higher is better. Each cell is colored to indicate the best performance in each column.}
    \label{tab:kitti_comparison}
    \resizebox{0.8\textwidth}{!}{%
    \begin{tabular}{l ccc ccc ccc}
        \toprule
        \multirow{2}{*}{\textbf{Method}} & \multicolumn{3}{c}{\textbf{Lateral Recall (\%)}} & \multicolumn{3}{c}{\textbf{Longitudinal Recall (\%)}} & \multicolumn{3}{c}{\textbf{Orientation Recall (\%)}} \\
        \cmidrule(lr){2-4} \cmidrule(lr){5-7} \cmidrule(lr){8-10}
         & 1m & 3m & 5m & 1m & 3m & 5m & 1$^{\circ}$ & 3$^{\circ}$ & 5$^{\circ}$ \\
        \midrule
        DSM             & 10.77 & 31.37 & 48.24 & 3.87 & 11.73 & 19.50 & 3.53 & 14.09 & 23.95 \\
        VIGOR           & 17.38 & 48.20 & 70.79 & 4.07 & 12.52 & 20.14 & -    & -     & -     \\
        BeyondRetrieval & 27.82 & 59.79 & 72.89 & 5.75 & 16.36 & 26.48 & 18.42 & 49.72 & 71.00 \\
        OrienterNet     & 51.26 & 84.77 & 91.81 & 22.39 & 46.79 & 57.81 & 20.41 & 52.24 & 73.53  \\
        \textbf{Ours}   & \textbf{65.95} & \textbf{90.08} & \textbf{94.67} & \textbf{23.51} & \textbf{51.72} & \textbf{63.03} & \textbf{26.00} & \textbf{66.07} & \textbf{84.27} \\
        \bottomrule
    \end{tabular}%
    }
\end{table*}

\begin{table*}[htbp]
    \centering
    \caption{\textbf{Quantitative comparison on the MGL and nuScenes datasets.} All metrics are Recall Accuracy (\%), where higher is better. The best performance for each column within each dataset group is highlighted.}
    \label{tab:mgl_nuscenes_comparison}
    \resizebox{0.8\textwidth}{!}{%
    \begin{tabular}{ll cccc cccc}
        \toprule
        \multirow{2}{*}{\textbf{Dataset}} & \multirow{2}{*}{\textbf{Method}} & \multicolumn{4}{c}{\textbf{Position Recall (\%)}} & \multicolumn{4}{c}{\textbf{Orientation Recall (\%)}} \\
        \cmidrule(lr){3-6} \cmidrule(lr){7-10}
        & & 1m & 2m & 5m & 10m & 1$^{\circ}$ & 2$^{\circ}$ & 5$^{\circ}$ & 10$^{\circ}$ \\
        \midrule
        \multirow{2}{*}{MGL} & OrienterNet & 10.78 & 29.88 & 54.72 & 67.25 & 18.98 & 35.15 & 63.03 & 76.63 \\
        & \textbf{Ours} & \textbf{11.07} & \textbf{31.46} & \textbf{57.23} & \textbf{69.30} & \textbf{19.57} & \textbf{35.74} & \textbf{64.79} & \textbf{77.91} \\
        \midrule
        \multirow{2}{*}{nuScenes} & OrienterNet & 2.89 & 6.01 & 18.57 & 38.49 & 9.30 & 16.86 & 35.40 & 55.81 \\
        & \textbf{Ours} & \textbf{15.70} & \textbf{28.83} & \textbf{56.74} & \textbf{79.20} & \textbf{19.32} & \textbf{37.46} & \textbf{70.41} & \textbf{86.91} \\
        \bottomrule
    \end{tabular}%
    }
\end{table*}

\begin{table}[htbp]
    \centering 
    \caption{\textbf{Ablation study of the Trajectory Refinement module on the KITTI dataset.} The metric is Position Recall Accuracy, where higher is better. The best results are highlighted.}
    \label{tab:ablation_refinement_condensed}
    \resizebox{0.45\textwidth}{!}{%
        \begin{tabular}{l cccc}
            \toprule
            \multirow{2}{*}{\textbf{Method}} & \multicolumn{4}{c}{\textbf{Position Recall}} \\
            \cmidrule(lr){2-5}
            & 1m & 2m & 5m & 10m \\
            \midrule
            w/o Trajectory Refinement & 0.0978 & 0.2871 & 0.5325 & 0.6591 \\
            \textbf{Ours} & \textbf{0.1554} & \textbf{0.3530} & \textbf{0.5935} & \textbf{0.7075} \\
            \bottomrule
        \end{tabular}%
    }
\end{table}

\subsection{Implementation Details}

\textbf{Input Representation.}
Our method follows the setup of previous work for a fair comparison. We use a single front-view image as the visual input. The map input is a 128m $\times$ 128m tile extracted from a rasterized navigation map, centered on the ego-vehicle's noisy GPS position. The map tile has a resolution of 0.5 meters per pixel (mpp).

\textbf{Training Setup.}
To simulate real-world GPS errors and train a robust model, we apply random perturbations to the ground truth pose for each training sample. These perturbations are sampled uniformly from a rotation range of $\theta \in [-30^{\circ}, 30^{\circ}]$ and a translation range of $t \in [-30\,\text{m}, 30\,\text{m}]$. Our model is trained end-to-end using the AdamW optimizer with a learning rate of $1 \times 10^{-4}$ and a weight decay of $1 \times 10^{-2}$. The implementation is based on PyTorch. We train the DiffVL model on a single NVIDIA RTX 2080 GPU.

\subsection{Datasets}
We conduct extensive experiments on three large-scale and diverse datasets: KITTI \cite{kitti}, MGL \cite{sarlin2023orienternet}, and nuScenes \cite{caesar2020nuscenes}. Below, we provide a brief introduction to each and highlight the specific challenges they pose.

\textbf{KITTI.}
The KITTI dataset \cite{kitti} is an authoritative benchmark widely used in the autonomous driving domain. Collected in and around Karlsruhe, Germany, it covers diverse real-world traffic scenarios, ranging from dense urban streets and rural roads to high-speed highways. Its challenging sequences with dynamic objects and varied lighting conditions make it ideal for evaluating localization robustness. KITTI provides precisely synchronized and calibrated multi-modal sensor data, accompanied by ground truth poses generated by a high-grade GPS/IMU system. We adhere to the official training and testing splits for our experiments.

\textbf{MGL.}
The MGL dataset was introduced by OrienterNet \cite{sarlin2023orienternet} to facilitate research in large-scale visual geo-localization. It was collected from the Mapillary platform and comprises over 760k images from 12 cities across Europe and the US. This dataset is particularly challenging due to its immense diversity; images were captured by various cameras (handheld, mounted on cars, bikes) under a wide range of weather and lighting conditions. This diversity tests the generalization capability of a model. All images are accompanied by ground truth (GT) poses and OpenStreetMap (OSM) data. As of this writing, data from two cities (Amsterdam and Vilnius) are no longer accessible. Therefore, we utilize the data from the remaining 10 cities for both training and evaluation.

\textbf{nuScenes.}
The nuScenes \cite{caesar2020nuscenes} dataset is a widely-used large-scale autonomous driving dataset, known for its comprehensive sensor suite and complex urban environments. It consists of 1,000 driving scenarios captured in Boston and Singapore, with each scene being 20s long. The dataset features dense traffic, complex intersections, and significant pedestrian activity, making it an excellent testbed for evaluating performance in safety-critical situations. We follow the official data splits, using the training set of 750 sequences (approx. 28,000 frames) and the validation set of 150 sequences (approx. 6,000 frames) for our experiments.

\subsection{Localization Results}

\textbf{Quantitative Analysis on KITTI.}
We first evaluate our method on the KITTI dataset, comparing it with state-of-the-art methods including OrienterNet\cite{sarlin2023orienternet}, DSM\cite{shi2020looking}, VIGOR\cite{zhu2021vigor}, and BeyondRetrieval\cite{shi2022beyond}. Following standard evaluation protocols, we use Lateral Recall@Xm, Longitudinal Recall@Xm, and Orientation Recall@X° as our primary metrics. As presented in Table 1, DiffVL significantly outperforms all baseline methods across every metric.

\textbf{Performance on Large-Scale MGL and nuScenes Datasets.}
To demonstrate the scalability and generalization of our approach, we conduct further comparisons on the MGL and nuScenes datasets. The evaluation metrics are Recall Accuracy (RA) at distance thresholds of {1, 2, 5, 10} meters and Orientation Recall Accuracy at thresholds of {1, 2, 5, 10} degrees. The experimental results, summarized in Table 2, confirm the superiority of our method. On the highly diverse MGL dataset, DiffVL's consistent lead suggests that our model learns a generalizable representation rather than overfitting to a specific city or camera type. On the complex urban nuScenes dataset, our method's strong performance underscores its robustness in dense traffic and perceptually challenging scenarios.

\textbf{Visualization of Localization Results.}
Figure \ref{visualize} visualizes the localization results of our method on the KITTI dataset. In these visualizations, the red trajectory represents the ground truth (GT) GPS trajectory from the dataset, while the blue trajectory is the noisy GPS trajectory we synthetically generate. Given the noisy blue trajectory and a single image as input, our method produces the refined green "Generated Location" trajectory.

As shown in the figure, our method's output trajectory closely aligns with the ground truth in the purple region (which corresponds to the location of the input image), achieving high-quality localization. It is worth noting that for GPS points outside the purple region, the accuracy of the generated trajectory shows a slight decrease. This is expected, as our inference process only utilizes the single image from the purple region; the visual information corresponding to the other GPS points is unknown, thus precluding accurate visual localization for those points. However, this demonstrates that in the region where visual information is available, our method can achieve high-quality visual localization performance.

\subsection{Ablation Study}
To isolate and verify the contribution of our core Trajectory Refinement module, we performed a targeted ablation study on the KITTI dataset. We configured a baseline variant, denoted as w/o Trajectory Refinement, by removing the diffusion head and its associated loss ($\mathcal{L}_{\text{diff}}$). This variant relies solely on the BEV-map matching mechanism for localization, similar to conventional approaches.

The results, summarized in Table 3, clearly showcase the module's critical role. Removing the trajectory refinement leads to a significant degradation in performance across all position recall metrics. It is the diffusion model's ability to denoise the temporal GPS sequence and impose a kinematically coherent prior that is essential for achieving high-precision results. This ablation provides compelling evidence that our proposed conditional denoising paradigm is the key driver of DiffVL's state-of-the-art performance.

\section{Conclusion}
In this paper, we introduced DiffVL, a novel framework that pioneers a new paradigm for visual localization by reformulating the task as a conditional GPS denoising problem. Our core contribution is to reframe noisy GPS signals as a valuable generative prior, providing a methodological basis for recovering high-precision poses using a diffusion model. We actualize this through a dual-objective training strategy that synergistically guides the model to produce kinematically coherent trajectories while simultaneously learning a geometrically-grounded representation via visual-to-map alignment. Extensive experiments on large-scale datasets, including KITTI, MGL, and nuScenes, validate our approach, demonstrating that DiffVL consistently achieves state-of-the-art performance. 


\bibliographystyle{main}
\bibliography{main}
\addtolength{\textheight}{-12cm} 

\end{document}